\pdfoutput=1

\documentclass[11pt]{article}

\usepackage{ACL2023}

\usepackage{times}
\usepackage{latexsym}
\usepackage{enumitem}
\usepackage{booktabs}
\usepackage{subfigure}  
\usepackage{multirow} 
\usepackage[T1]{fontenc}

\usepackage[utf8]{inputenc}
\usepackage{microtype}
\usepackage{inconsolata}
\usepackage{graphicx}
\usepackage{tabularx}
\usepackage{enumitem}
\usepackage{amsmath}

\title{Health-LLM: Personalized Retrieval-Augmented Disease Prediction System}

\author{
    Qinkai Yu$^{5}$, Mingyu Jin$^{1}$, Dong Shu$^{3}$, Chong Zhang$^{2}$, Lizhou Fan$^4$, Wenyue Hua$^1$,\\\textbf{Suiyuan Zhu$^{7}$}, \textbf{Yanda Meng$^{5}$}, \textbf{Zhenting Wang}$^1$, \textbf{Mengnan Du$^6$},  \textbf{Yongfeng Zhang$^1$}\\ \\
    $^1$ Rutgers University, ~ $^2$ University of Liverpool,~ $^3$ Northwestern University, ~\\ $^4$ University of Michigan, ~$^5$ University of Exeter, ~ $^6$ New Jersey Institute of Technology, \\~ $^7$ New York University 
}

\begin{document}

\maketitle



\begin{abstract}
Recent advancements in artificial intelligence (AI), especially large language models (LLMs), have significantly advanced healthcare applications and demonstrated potential in intelligent medical treatment. To promote professional and personalized healthcare, we propose an innovative framework, \textbf{Health-LLM}, which combines large-scale feature extraction and trade-off scoring of medical knowledge. Compared to traditional health management applications, our system has three main advantages: (1) It integrates health reports and medical knowledge into a large model to ask relevant questions to the Large Language Model for disease prediction; (2) It leverages a retrieval augmented generation (RAG) mechanism to enhance feature extraction; (3) It incorporates a semi-automated feature updating framework that can merge and delete features to improve the accuracy of disease prediction. We experimented with a large number of health reports to assess the effectiveness of the Health-LLM system. The results indicate that the proposed system surpasses the existing ones and has the potential to advance disease prediction and personalized health management significantly. 

\end{abstract}

\section{Introduction}
\begin{figure*}[ht]
    \centering
    \includegraphics[scale=0.45]{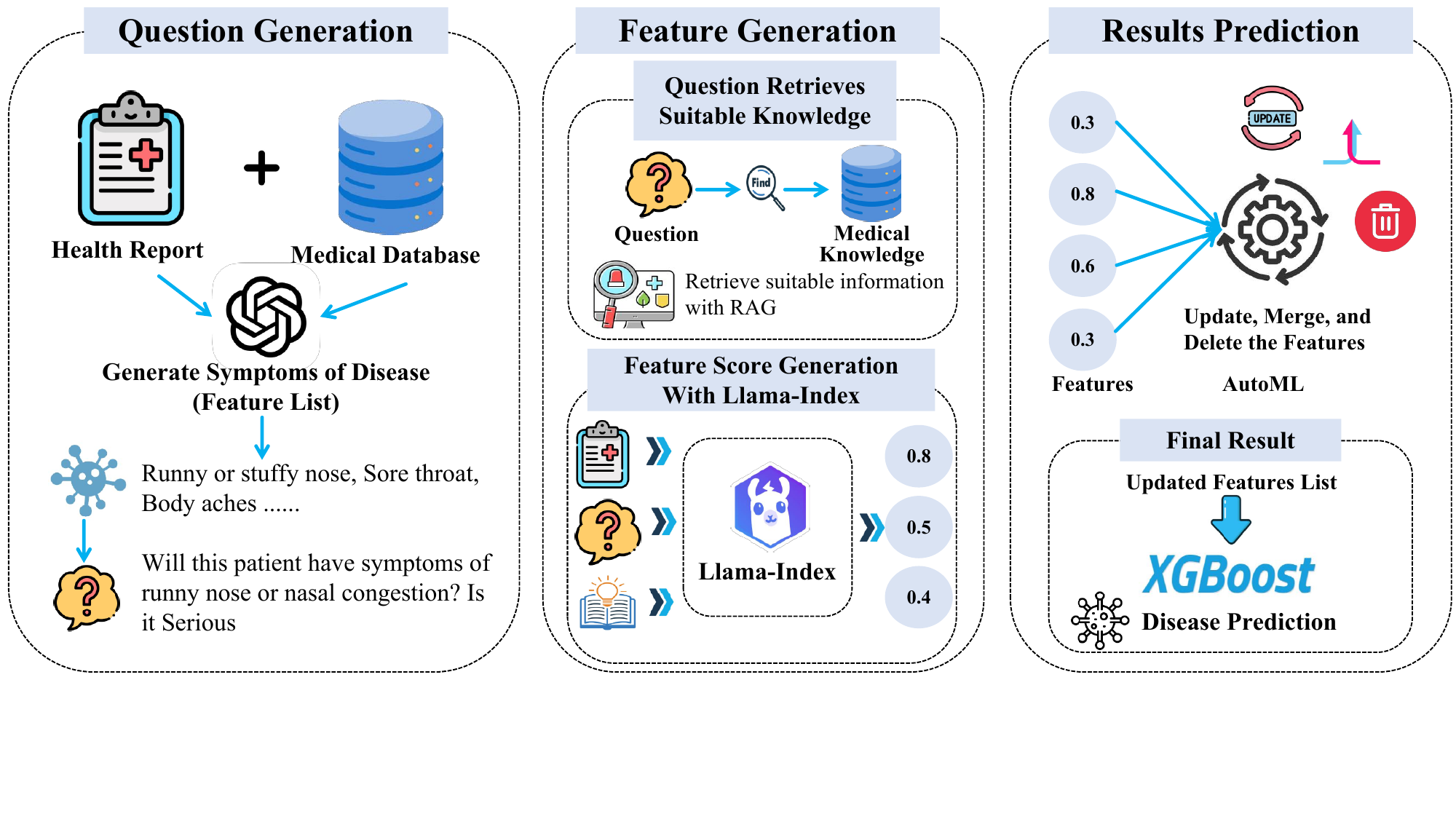}
    \caption{The overall workflow of the Health-LLM: from feature extraction to XGBoost prediction.}
    \vspace{-15pt}
    \label{fig:ICL-1}
\end{figure*}

The integration of AI into healthcare, notably through large language models (LLMs) such as GPT-3.5 \cite{rasmy2021med} and GPT-4~\cite{openai2023gpt4}, has reshaped the field of health management. Recent studies highlight the crucial role of LLM in using machine learning to improve healthcare outcomes~\cite{biswas2023role,singhal2022large}. Advancements in AI for healthcare demonstrate a shift towards models that handle complex medical data and offer improved precision. Nonetheless, traditional health management methods often struggle with the constraints imposed by static data and uniform standards, making them unfit to fully meet individual needs \cite{uddin2019comparing,lopez2020case,beam2018big,ghassemi2020review}. The health reports of patients offer a wealth of data, this information has the potential to predict future health issues and tailor health recommendations, but the difficulty lies in transforming these extensive data into practical insights.

This study focuses on the Clinical Prediction with Large Language Models (CPLLM) approach \cite{shoham2023cpllm}, which showcases the superior predictive capabilities of LLMs fine-tuned on clinical data. In particular, we propose an innovative system, \textbf{Health-LLM}, utilizing data analytics, machine learning, and medical knowledge for comprehensive health management. The system can provide users with personalized health recommendations based on predicted health risks (see \autoref{fig:ICL-1}), ultimately helping prevent future health complications. 

Specifically, the system uses the Llama Index \cite{liu8llamaindex} framework to analyze the information from the patient's health report. Then it assigns different scores to these features by the Llama Index, which is prepared with professional medical information by RAG. The scoring method asks the language model questions about the patient's condition. Our system also incorporates automated feature engineer technology \cite{he2021automl} to perform iterative optimization to extract important features and stable weights and scores. Finally, the system is trained based on the XGBoost model~\cite{chen2016xgboost} to make early predictions of existing diseases and provide personalized health recommendations to individuals.

We compare the performance of our system with traditional methods (including Pretrain-BERT \cite{devlin2018bert}, TextCNN \cite{chen2015convolutional}, Hierarchical Attention \cite{yang2016hierarchical}, Text BiLSTM with Attention \cite{liu2019bidirectional}, RoBERT \cite{liu2019roberta}) as well as mainstream large-scale language models(GPT-3.5, GPT-4, LLaMA-2 \cite{touvron2023llama}) in three different settings (zero-shot, few-shot, and information retrieval) to show the effectiveness of our system. Among them, the accuracy of GPT-4 combined with information retrieval by retrieval augmented generation (RAG) for disease diagnosis is 0.68, and the F1 score is 0.71, while our system has achieved an accuracy of 0.833 and an F1 score of 0.762, respectively.
Our key contributions are as follows:
\begin{itemize}[leftmargin=*]\setlength\itemsep{-0.3em}
\item We propose an innovative Health-LLM framework that combines large-scale feature extraction, precise scoring of medical knowledge using the Llama Index structure, and machine learning techniques to enable personalized disease prediction from patient health reports.
\item Our proposed Health-LLM framework achieves state-of-the-art performance on disease prediction tasks, surpassing existing methods like GPT-4 and fine-tuned LLaMA-2 models as demonstrated through extensive experiments.
\end{itemize}


\section{Background}

\subsection{AI for Health Management}
AI is revolutionizing healthcare through machine learning and relevant methods to enhance healthcare outcomes. This evolution is significantly driven by the emergence of LLMs, as seen in studies such as \cite{biswas2023role,singhal2022large,rasmy2021med}. These models are vital in clinical applications, including disease prediction and diagnosis. The intersection of AI and healthcare has seen notable progress, fueled by the availability of extensive health datasets and the advancement of sophisticated LLMs. Recent research, such as \cite{wang2023coad}, demonstrates the immense potential of LLMs in the healthcare sector, where they are used to understand and generate health reports and evaluate various health situations. 

A key development in this field is the Clinical Prediction with Large Language Models (CPLLM), which highlights the potential of LLMs fine-tuned on clinical data \cite{shoham2023cpllm}. CPLLM, using historical diagnosis records, has shown superiority over traditional models such as logistic regression and even advanced models such as Med-BERT \cite{rasmy2021med} in predicting future disease diagnoses. Another significant advancement in AI for health is the COAD framework \cite{wang2023coad}, which addresses the limitations of previous Transformer-based automatic diagnosis methods. AMIE (Articulate Medical Intelligence Explorer) is a medical knowledge graph created by Google \cite{tu2024towards}. It extracts and stores medical knowledge, including information about diseases, symptoms, and treatments. These advancements indicate a trend in AI for health, shifting towards models that effectively manage the complexity and subtleties of medical data. The progress made by CPLLM, COAD and AMIE underscores the transformative impact these technologies can have on healthcare, enhancing precision, efficiency, and personalization in patient care.
\subsection{Retrieval Augmented Generation}
The retrieval augmented generation (RAG) \cite{lewis2020retrieval} method is a natural language processing model that combines retrieval and generation components to handle knowledge-intensive tasks. The method consists of two stages: Retrieval and Generation. During the retrieval phase, RAG employs the Dense Passage Retrieval (DPR) system to retrieve the most relevant documents from a large-scale document database that answers the input question. The input question is encoded as a vector, which is then compared with the document vectors in the database to locate the most relevant documents. The main idea behind this method is to use a large-scale document collection to enhance the generation model's ability and improve the model's efficiency in dealing with complex and knowledge-dependent problems.



\section{Implementation}


\subsection{In-context Learning for Symptom Features Generation}

\label{sec:icl}

\begin{figure}[t]
    \centering
    \includegraphics[scale=0.75]{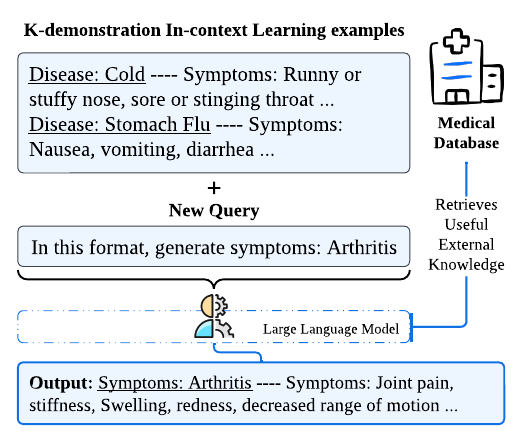}
    \caption{In-context learning workflow of symptom features generation in Health-LLM.}
    \label{fig:ICL-2}
    \vspace{-0.3cm}
\end{figure}

In the initial phase of our implementation, we systematically extract symptom features from a range of diseases by harnessing the in-context learning capabilities of LLMs. \autoref{fig:ICL-2} shows an example of the corresponding workflow. We first prompt the model with a series of examples, such as "disease: cold, symptoms: runny or stuffy nose, sore or tingling throat, cough, sneeze", to teach it the pattern generated by the symptom profile. Leveraging this in-context learning paradigm, our system is ready to take new query inputs and efficiently produce symptom descriptors for various diseases in a batch processing mode.

To enhance the precision of this generative process, we integrate a supplementary medical knowledge base, employing a RAG mechanism for enriched knowledge retrieval. Through the guidance of the system prompts, the LLM is asked to answer questions about the extracted symptoms. Since LLM may lack specialized knowledge in medicine, we provide contextual information for these questions in an embedded form. Thus, we utilize advanced RAG technology to synchronize our queries with the knowledge base. In particular, RAG helps identify and retrieve the three most relevant pieces of information that align with the symptoms mentioned in the question. Our system then extracts these pieces of information and seamlessly integrates them to enrich the input prompts for the models.

\subsection{Assigning Score by Llama Index}
\label{sec:assign}


To integrate LLMs from different sources, we adopt the Llama Index framework \cite{liu8llamaindex} for question answering (QA) in health reports. This approach allows us to take full advantage of advanced natural language processing models to extract features, a key step in our ability to predict disease and provide health advice. 

To ensure accurate and domain-specific responses, we then prompt the system with our high-quality generated queries relevant to the health issue. The LLM then assigns a confidence score between 0 and 1, indicating the system's perception of the health issue's severity level. For instance, the question "Does this person have good sleeping habits?" might receive a response like "Sleep: 0.6," suggesting moderately positive sleeping patterns. This numerical confidence score then becomes a key attribute in our classification model.

  



The Llama index serves to streamline document-based QA through a strategic "search-then-synthesize" approach. The process unfolds as follows. Initially, health report documents are curated and formatted into plain text. These documents are then segmented into smaller manageable text blocks. Each block is processed through a text-embedding interface, transforming it into a vector representation that is subsequently stored within a vector database; here, OpenAI’s embeddings can be utilized for this transformation \cite{openai_embeddings}.

When receiving questions, the system converts them into vectors to facilitate search within the vector database, aiming to identify the most relevant text block(s). The identified text block is then amalgamated with the query to formulate a refined request. This newly created request is sent to the OpenAI API for processing. In total, we create a list of 152 questions and an area knowledge database. All the scores resulting from answering these questions will be used as features to enter the downstream machine learning model. For example: Sleep: 0.6, which means that the person's sleep condition is okay.

\begin{table*}[t!]
\caption{An example of the interactive process in case study.}
\small
	\centering
	\begin{tabularx}{1\textwidth}{lX}
		\toprule
		 & \textbf{User Inputs and Conversation with Health-LLM}  \\
		\midrule
User: & In the morning, the stools are normal and thin, and they are a little beaded at night.\\
Health-LLM:& Hello, how long has this situation? Is the diet regular? Can you provide me with basic information such as your age and gender? What is the feeling of abdomen? \\
User: & It has been four or five days. I have a stool when I ate Amoshilin in the first rhinitis. After a few days of laxative spirit, it is like this now. Haven't eaten food and cold.\\
		\midrule
& \textbf{Prediction and Personal Advice}\\
		\midrule
Health-LLM:& Gastrointestinal dysfunction and Diarrhea. Choose light meals and avoid stimulating foods. You don't need to take medicine. You are welcome. Wish you health and happiness!\\
		\bottomrule
	\end{tabularx}%
  \label{tab:33}
\vspace{-11pt}
\end{table*}
\subsection{Predictive Model and Health Advice}
\label{sec:prediction}
In our quest to develop a robust disease classification system, we have established a comprehensive framework that encompasses 61 disease labels, ranging from common ailments such as insomnia and indigestion to more complex conditions like endocrine disorders. We fit the features with XGBoost, which allows the XGBoost model to fit the feature scores extracted from the Llama index, and learn the feature representation of each disease under the Llama index. The results of XGBoost fitting in this framework are multiple binary combinations, with "0" indicating no associated disease and "1" indicating an associated disease. In addition, we have also classified certain diseases (e.g., fatty liver) at a finer level of granularity. For example, "0" indicates mild fatty liver and "1" indicates severe fatty liver.

The significance of domain-specific knowledge incorporation has become clear in our quest to improve disease prediction through feature preprocessing. Addressing this, we use Context-Aware Automated Feature Engineering (CAAFE), which utilizes the power of LLMs \cite{hollmann2023gpt} to generate features iteratively with semantic relevance informed by the dataset’s context. We have evolved from a semi-automated system to a fully automated one. We now use LLMs to autonomously craft feature and dataset descriptions, thereby streamlining the feature engineering process and enriching our models with contextually meaningful data. After the prediction is completed, we will query the LLM for the last time with the predicted diseases, and let it generate targeted health suggestions based on the health report and professional knowledge.  
\subsection{Interact with Health-LLM}
Users have two ways to interact with the Health-LLM system. First, the user can submit his/her health report to the system and get the prediction and the health advice. Second, the user can describe his/her health problem and ask the system, the system will record their dialog and save as a health dialog. The dialog messages are similar to the health reports in the training process. Health-LLM generates the corresponding disease predictions and gives personal suggestions based on the dialog messages following the process shown in \autoref{fig:ICL-1}. The large language model responsible for the dialog and giving personal suggestions in Health-LLM is GPT-4 turbo. In brief, Health-LLM provides diagnosis and personalized advice through interaction with the user.

\section{A Case Study}
In this section, we describe the entire process by which our system accurately predicts a patient's disease and compare its performance with, for example, using a large model alone. Our case is under dataset IMCS-21 (Table 1) \cite{chen2023benchmark}.
\noindent \textbf{Settings.} 
To our knowledge, IMCS-21 is currently the only dataset on telemedicine consultation in China. The data set contains a total of 4,116 annotated samples with 164,731 utterances across 10 pediatric diseases, including \textit{bronchitis}, \textit{fever}, \textit{diarrhoea}, \textit{upper respiratory infection}, \textit{dyspepsia}, \textit{cold}, \textit{cough}, \textit{jaundice}, \textit{constipation} and \textit{bronchopneumonia}. Each dialogue contains an average of 40 utterances, 523 Chinese characters (580 characters if including self-report) and 26 entities. To prevent data leakage, we pre-processed the data by using GPT4 to eliminate content containing specific medical conditions from the data conversations and transformed the conversations into the form of an Electronic Patient Record (EPR). We use two metrics: ACC(Accuracy), F1(Macro F1-score), to evaluate the prediction quality.

\subsection{Health-LLM Diagnostic Test and Comparative Experiment}


\noindent \textbf{Health-LLM Diagnostic Test.} 
Trained on the given dataset IMCS-21, our system is capable of taking complex patient inputs and provide medical suggestions. We provide an example of the user-system interaction in \autoref{wide_table2} in the Appendix. To employ this system, the user inputs their information such as physical conditions and disease symptoms. The system predicts potential diseases for the user based on prior experience and provides the appropriate diagnosis, as well as follow-up recommendations. We give an example in \autoref{tab:33}. Our system plays the role of a doctor who guides the user step by step in describing his or her real symptoms. The output of the system is the specific name of the disease and reasonable suggestions. Our validation process uses conversations from the validation set as input to simulate user adoption.

\noindent \textbf{Comparative Experiment.}
\begin{table}[t]
    \centering
    \renewcommand\arraystretch{0.85}

    \caption{Comparing with existing methods on the Health-LLM diagnostic test.}
    \resizebox{0.5\textwidth}{!}{
    \begin{tabular}{l|cccc}
    \hline \textbf{Models} & \textbf{Accuracy}& \textbf{F1}\\
    \hline 
           Pretrain-Bert with XGBoost & 0.326 & 0.303\\
           GPT-3.5(zero-shot)& 0.333& 0.361 \\
           GPT-3.5(few shot) & 0.381 & 0.349\\
           GPT-4(zero-shot)& 0.390& 0.312 \\
           TextCNN & 0.437 & 0.429\\
           GPT-3.5(few shot with information retrieval) & 0.451& 0.451 \\
           Hierarchical Attention & 0.495 & 0.477\\
           Text BiLSTM with Attention & 0.512  & 0.500\\
           RoBERT & 0.585  & 0.543\\
           GPT-4(few shot) & 0.620& 0.671 \\
           GPT-4(few shot with information retrieval) & 0.680& 0.718 \\
           Fintuned-LLaMA-2-7B & 0.710 & 0.593 \\
           Fintuned-LLaMA-2-13B & 0.730 & 0.671\\
           \hline
    \textbf{Health-LLM (Ours)} & \textbf{0.833}& \textbf{0.762} \\
    \hline
    \end{tabular} }
    \vspace{-12pt}
    \label{tab:comp_medQA}
\end{table}
Our comparative analysis involves two open-source models, GPT-3.5 and GPT-4. We deploy these models in a zero-shot setting to diagnose cases.  Following this, we will transition to a few-shot context, providing each model with a selection of prior predictions to enhance its diagnostic capabilities. Subsequently, we will integrate supplementary medical knowledge into the larger models to refine their predictions further. The culmination of this process will involve leveraging the dataset we use in Health-LLM to fine-tune LLaMA-2 \cite{touvron2023llama}, with the objective of observing the impact on its performance. For extended texts and chapter-level analysis in the traditional method, researchers have introduced a framework known as Hierarchical Attention \cite{yang2016hierarchical} and RoBERTa \cite{liu2019roberta}, which excels in handling multi-classification challenges in case texts. The conventional approach fails in accurately diagnosing diseases from medical reports, as it lacks the deep text comprehension exhibited by larger models. Additionally, it struggles to encapsulate all the pertinent information within lengthy paragraphs. This approach ranks impressively, just behind advanced models like GPT-4 and fine-tuned Llama-2 in terms of performance. The comparative test results are shown in \autoref{tab:comp_medQA}. Experiments showed that our system achieved an overall prediction accuracy of 83\% for the disease, with F1 score of 0.762. Health-LLM achieves the best performance compared to direct inference using other large models. 
\textbf{Ablation Study.} In this section, we verify the effectiveness of each component of our Health-LLM. The results are displayed in \autoref{tab:ab_medk}. The experimental results have two implications. First, it is necessary to index professional healthcare data to ensure the accuracy of diagnostic reasoning. Our diagnostic reasoning accuracy has improved significantly in the group with indexed professional healthcare knowledge. Second, our data processing and feature extraction with CAAFE is effective.
\begin{table}[ht]
    \centering
        \caption{The result of Health-LLM diagnostic test on IMCS-21 dataset.}
        \begin{tabular}{l|cccc}
            \hline 
            \textbf{Model}  & \textbf{Acc} & \textbf{F1}\\
            \hline 
            Health-LLM without Retrieval& 0.78 & 0.714\\
            Health-LLM without CAAFE &  0.77& 0.721\\

            Health-LLM&  0.83& 0.762\\
            \hline
        \end{tabular}
        \vspace{-10pt}
    \label{tab:ab_medk}
\end{table}
\section{Conclusions}
We present a novel system, Health-LLM, that combines large-scale feature extraction, precise scoring of medical knowledge, and machine learning techniques to make better use of patient health reports. It improves the prediction of potential diseases compared to GPT-3.5, GPT-4, and finetuned LLaMA-2. Our system is capable to predict potential diseases and provide customized health advice to individuals. In addition to its practical advancements, Health-LLM also serves as a proof-of-concept LLM-based system in healthcare, highlighting the huge potential of further development of AI applications in the health field.

%


\newpage

\bibliography{anthology,custom}

\begin{thebibliography}{24}
\expandafter\ifx\csname natexlab\endcsname\relax\def\natexlab#1{#1}\fi

\bibitem[{Achiam et~al.(2023)Achiam, Adler, Agarwal, Ahmad, Akkaya, Aleman, Almeida, Altenschmidt, Altman, Anadkat et~al.}]{openai2023gpt4}
Josh Achiam, Steven Adler, Sandhini Agarwal, Lama Ahmad, Ilge Akkaya, Florencia~Leoni Aleman, Diogo Almeida, Janko Altenschmidt, Sam Altman, Shyamal Anadkat, et~al. 2023.
\newblock Gpt-4 technical report.
\newblock \emph{arXiv preprint arXiv:2303.08774}.

\bibitem[{Beam and Kohane(2018)}]{beam2018big}
Andrew~L Beam and Isaac~S Kohane. 2018.
\newblock Big data and machine learning in health care.
\newblock \emph{Jama}, 319(13):1317--1318.

\bibitem[{Biswas(2023)}]{biswas2023role}
Som~S Biswas. 2023.
\newblock Role of chat gpt in public health.
\newblock \emph{Annals of biomedical engineering}, 51(5):868--869.

\bibitem[{Chen and Guestrin(2016)}]{chen2016xgboost}
Tianqi Chen and Carlos Guestrin. 2016.
\newblock Xgboost: A scalable tree boosting system.
\newblock In \emph{Proceedings of the 22nd acm sigkdd international conference on knowledge discovery and data mining}, pages 785--794.

\bibitem[{Chen et~al.(2023)Chen, Li, Fang, Yao, Zhong, Hao, Zhang, Huang, Peng, and Wei}]{chen2023benchmark}
Wei Chen, Zhiwei Li, Hongyi Fang, Qianyuan Yao, Cheng Zhong, Jianye Hao, Qi~Zhang, Xuanjing Huang, Jiajie Peng, and Zhongyu Wei. 2023.
\newblock A benchmark for automatic medical consultation system: frameworks, tasks and datasets.
\newblock \emph{Bioinformatics}, 39(1):btac817.

\bibitem[{Chen(2015)}]{chen2015convolutional}
Yahui Chen. 2015.
\newblock Convolutional neural network for sentence classification.
\newblock Master's thesis, University of Waterloo.

\bibitem[{Devlin et~al.(2018)Devlin, Chang, Lee, and Toutanova}]{devlin2018bert}
Jacob Devlin, Ming-Wei Chang, Kenton Lee, and Kristina Toutanova. 2018.
\newblock Bert: Pre-training of deep bidirectional transformers for language understanding.
\newblock \emph{arXiv preprint arXiv:1810.04805}.

\bibitem[{Ghassemi et~al.(2020)Ghassemi, Naumann, Schulam, Beam, Chen, and Ranganath}]{ghassemi2020review}
Marzyeh Ghassemi, Tristan Naumann, Peter Schulam, Andrew~L Beam, Irene~Y Chen, and Rajesh Ranganath. 2020.
\newblock A review of challenges and opportunities in machine learning for health.
\newblock \emph{AMIA Summits on Translational Science Proceedings}, 2020:191.

\bibitem[{He et~al.(2021)He, Zhao, and Chu}]{he2021automl}
Xin He, Kaiyong Zhao, and Xiaowen Chu. 2021.
\newblock Automl: A survey of the state-of-the-art.
\newblock \emph{Knowledge-Based Systems}, 212:106622.

\bibitem[{Hollmann et~al.(2023)Hollmann, M{\"u}ller, and Hutter}]{hollmann2023gpt}
Noah Hollmann, Samuel M{\"u}ller, and Frank Hutter. 2023.
\newblock Gpt for semi-automated data science: Introducing caafe for context-aware automated feature engineering.
\newblock \emph{arXiv preprint arXiv:2305.03403}.

\bibitem[{Lewis et~al.(2020)Lewis, Perez, Piktus, Petroni, Karpukhin, Goyal, K{\"u}ttler, Lewis, Yih, Rockt{\"a}schel et~al.}]{lewis2020retrieval}
Patrick Lewis, Ethan Perez, Aleksandra Piktus, Fabio Petroni, Vladimir Karpukhin, Naman Goyal, Heinrich K{\"u}ttler, Mike Lewis, Wen-tau Yih, Tim Rockt{\"a}schel, et~al. 2020.
\newblock Retrieval-augmented generation for knowledge-intensive nlp tasks.
\newblock \emph{Advances in Neural Information Processing Systems}, 33:9459--9474.

\bibitem[{Liu and Guo(2019)}]{liu2019bidirectional}
Gang Liu and Jiabao Guo. 2019.
\newblock Bidirectional lstm with attention mechanism and convolutional layer for text classification.
\newblock \emph{Neurocomputing}, 337:325--338.

\bibitem[{Liu(2022)}]{liu8llamaindex}
Jerry Liu. 2022.
\newblock Llamaindex, 11 2022.
\newblock \emph{URL https://github. com/jerryjliu/llama\_index.(Cited on page 8)}.

\bibitem[{Liu et~al.(2019)Liu, Ott, Goyal, Du, Joshi, Chen, Levy, Lewis, Zettlemoyer, and Stoyanov}]{liu2019roberta}
Yinhan Liu, Myle Ott, Naman Goyal, Jingfei Du, Mandar Joshi, Danqi Chen, Omer Levy, Mike Lewis, Luke Zettlemoyer, and Veselin Stoyanov. 2019.
\newblock \href {http://arxiv.org/abs/1907.11692} {Roberta: A robustly optimized bert pretraining approach}.

\bibitem[{L{\'o}pez-Mart{\'\i}nez et~al.(2020)L{\'o}pez-Mart{\'\i}nez, N{\'u}{\~n}ez-Valdez, Garc{\'\i}a-D{\'\i}az, and Bursac}]{lopez2020case}
Fernando L{\'o}pez-Mart{\'\i}nez, Edward~Rolando N{\'u}{\~n}ez-Valdez, Vicente Garc{\'\i}a-D{\'\i}az, and Zoran Bursac. 2020.
\newblock A case study for a big data and machine learning platform to improve medical decision support in population health management.
\newblock \emph{Algorithms}, 13(4):102.

\bibitem[{{OpenAI}(2021)}]{openai_embeddings}
{OpenAI}. 2021.
\newblock \href {https://platform.openai.com/docs/guides/embeddings/embedding-models} {Embedding models}.

\bibitem[{Rasmy et~al.(2021)Rasmy, Xiang, Xie, Tao, and Zhi}]{rasmy2021med}
Laila Rasmy, Yang Xiang, Ziqian Xie, Cui Tao, and Degui Zhi. 2021.
\newblock Med-bert: pretrained contextualized embeddings on large-scale structured electronic health records for disease prediction.
\newblock \emph{NPJ digital medicine}, 4(1):86.

\bibitem[{Shoham and Rappoport(2023)}]{shoham2023cpllm}
Ofir~Ben Shoham and Nadav Rappoport. 2023.
\newblock \href {http://arxiv.org/abs/2309.11295} {Cpllm: Clinical prediction with large language models}.

\bibitem[{Singhal et~al.(2022)Singhal, Azizi, Tu, Mahdavi, Wei, Chung, Scales, Tanwani, Cole-Lewis, Pfohl et~al.}]{singhal2022large}
Karan Singhal, Shekoofeh Azizi, Tao Tu, S~Sara Mahdavi, Jason Wei, Hyung~Won Chung, Nathan Scales, Ajay Tanwani, Heather Cole-Lewis, Stephen Pfohl, et~al. 2022.
\newblock Large language models encode clinical knowledge.
\newblock \emph{arXiv preprint arXiv:2212.13138}.

\bibitem[{Touvron et~al.(2023)Touvron, Martin, Stone, Albert, Almahairi, Babaei, Bashlykov, Batra, Bhargava, Bhosale et~al.}]{touvron2023llama}
Hugo Touvron, Louis Martin, Kevin Stone, Peter Albert, Amjad Almahairi, Yasmine Babaei, Nikolay Bashlykov, Soumya Batra, Prajjwal Bhargava, Shruti Bhosale, et~al. 2023.
\newblock Llama 2: Open foundation and fine-tuned chat models.
\newblock \emph{arXiv preprint arXiv:2307.09288}.

\bibitem[{Tu et~al.(2024)Tu, Palepu, Schaekermann, Saab, Freyberg, Tanno, Wang, Li, Amin, Tomasev et~al.}]{tu2024towards}
Tao Tu, Anil Palepu, Mike Schaekermann, Khaled Saab, Jan Freyberg, Ryutaro Tanno, Amy Wang, Brenna Li, Mohamed Amin, Nenad Tomasev, et~al. 2024.
\newblock Towards conversational diagnostic ai.
\newblock \emph{arXiv preprint arXiv:2401.05654}.

\bibitem[{Uddin et~al.(2019)Uddin, Khan, Hossain, and Moni}]{uddin2019comparing}
Shahadat Uddin, Arif Khan, Md~Ekramul Hossain, and Mohammad~Ali Moni. 2019.
\newblock Comparing different supervised machine learning algorithms for disease prediction.
\newblock \emph{BMC medical informatics and decision making}, 19(1):1--16.

\bibitem[{Wang et~al.(2023)Wang, Kwan, Wong, and Zheng}]{wang2023coad}
Huimin Wang, Wai-Chung Kwan, Kam-Fai Wong, and Yefeng Zheng. 2023.
\newblock Coad: Automatic diagnosis through symptom and disease collaborative generation.
\newblock \emph{arXiv preprint arXiv:2307.08290}.

\bibitem[{Yang et~al.(2016)Yang, Yang, Dyer, He, Smola, and Hovy}]{yang2016hierarchical}
Zichao Yang, Diyi Yang, Chris Dyer, Xiaodong He, Alex Smola, and Eduard Hovy. 2016.
\newblock Hierarchical attention networks for document classification.
\newblock In \emph{Proceedings of the 2016 conference of the North American chapter of the association for computational linguistics: human language technologies}, pages 1480--1489.

\end{thebibliography}
\bibliographystyle{acl_natbib}
\newpage
\onecolumn
\section*{Appendix}

\begin{table*}[h]
\small
  \centering
  \caption{Example of one health report we have made by dataset IMCS-21}
  \resizebox{1.0\textwidth}{!}{%
  \begin{tabular}{l}
    \toprule
    \multicolumn{1}{c}{\textbf{IMCS-21}} \\\midrule
    -Hello, there is a pain around the navel,
     I don’t know what's going on (female, 29 years old)\\
    -Hello, how long has this situation?\\
    -Two or three days.\\
    -It hurts, and it will not hurt for a while.\\
    -Have you used any medicine? Have you ever done?\\
    -There is no medication and no examination.\\
    -Is the stool normal?\\
    -normal.\\
    -Are there any other symptoms? Do you want to vomit?\\
    -No.\\
    -Is it faint pain?\\
    -Once appetite, a little bit bloated.\\
    -It may be gastrointestinal dysfunction.\\
    -Yes, faint pain.\\
    -Eat some song Meibing Try.relative.\\
    -It felt like a needle was tied, and it was fine for a few seconds.\\
    -It feels that the problem is not particularly big.\\
    -Try it if you take the medicine I said.\\
    -Is there a compound fairy crane grassyitis tablet at home?\\
    -This is mainly to treat diarrhea. Don't eat without diarrhea. Oh, alright.\\
    -Uh-huh.\\
    -Which aspect can cause gastrointestinal function?\\
    -It may be that the diet may be a mental factor or the autoimmune system.\\
    -It may also be the digestive system or autoimmune system. relative.\\
    -OK\\
    -Uh-huh.\\
    -It seemed a bit like a diarrhea. After eating 1 at noon, I wanted to pull it after a 
     while, and I was a little bit pulled.\\
    -You can eat the medicine you said.\\
    -Oh well.\\
    -Uh-huh.\\
    \bottomrule
    \end{tabular}}
  \label{wide_table2}
\end{table*}

\end{document}